\documentclass[lettersize,journal]{IEEEtran}
\usepackage{amsmath,amsfonts}
\usepackage[linesnumbered,ruled,vlined]{algorithm2e}
\usepackage{enumerate}
\usepackage{array, multirow}
\usepackage[caption=false,font=scriptsize,labelfont=sf,textfont=sf]{subfig}
\usepackage{textcomp}
\usepackage{stfloats}
\usepackage{hyperref}
\usepackage{verbatim}
\usepackage{graphicx}
\usepackage{cite}
\begin{document}

\title{Call-Center Staff Scheduling Considering Performance Evolution under Emotional Stress}

\author{Yu-Jun Zheng,~\IEEEmembership{Senior Member,~IEEE,}
	Xin-Ya Chen,
	Xue-Qin Lu,
        Wei-Guo Sheng, \IEEEmembership{Member,~IEEE,}
        and Sheng-Yong Chen, \IEEEmembership{Senior Member,~IEEE}
\thanks{Manuscript received ...(date to be filled in by Editor).  This work was supported by National Natural Science Foundation of China under Grant 61872123 and Grant 62372148.}
\thanks{Y.-J.~Zheng, Xin-Ya Chen, X.-Q. Lu, and W.-G. Sheng are with the School of Information Science and Technology, Hangzhou Normal University, Hangzhou 311121, China.}
\thanks{S.-Y. Chen is with School of Computer Science and Engineering, Tianjin University of Technology, Tianjin 300384, China.}
\thanks{This work has been submitted to the IEEE for possible publication. Copyright may be transferred without notice, after which this version may no longer be accessible.}}

\markboth{IEEE TRANSACTIONS ON CYBERNETICS,~Vol.~XX, No.~X, XX~2025}%
{Zheng \MakeLowercase{\textit{et al.}}: Call-Center Staff Scheduling Considering Performance Evolution under Emotional Stress}


\maketitle

\begin{abstract}
Emotional stress often has a significant effect on the working performance of staff, but this effect is commonly neglected in existing staff scheduling methods. We study a call-center staff scheduling problem, which considers the evolution of work performance of staff under emotional stress. First, we present an emotional stress driven model that estimates the working performance of call-center employees based on not only skill levels but also emotional states. On the basis of the model, we formulate a combined short-term and long-term call-center staff scheduling problem aiming at maximizing the customer service level, which depends on the working performance of employees. We then propose a memetic optimization algorithm combining global mutation and neighborhood search assisted by deep reinforcement learning to efficiently solve this problem. Experimental results on real-world problem instances of bank call-center staff scheduling demonstrate the performance advantages of the proposed method over selected popular staff scheduling methods. By explicitly modeling and incorporating emotional stress, our method reflects a more realistic understanding and utilization of human behavior in staff scheduling.
\end{abstract}


\section{Introduction}
\IEEEPARstart{C}{all}-center staff scheduling, which involves determining the number of staff and assigning them with service requests (tasks), is a critical aspect of running a service-oriented organization smoothly. It belongs to the more general category of staff scheduling (also known as personnel scheduling or rostering), which has been studied in a variety of application areas \cite{Ernst04EJOR} such as airline, health, and finance. For organizations with a large workforce and/or high service loads, developing efficient solutions is computationally difficult, as the problem typically involves various complex objectives and constraints, including improving customer satisfaction, reducing costs and other resources, meeting individual preferences, achieving fairness among employees, satisfying laws/regulations, etc \cite{Nguyen22TCyb}.

\begin{figure}[!t]
\centering\includegraphics[scale=0.63]{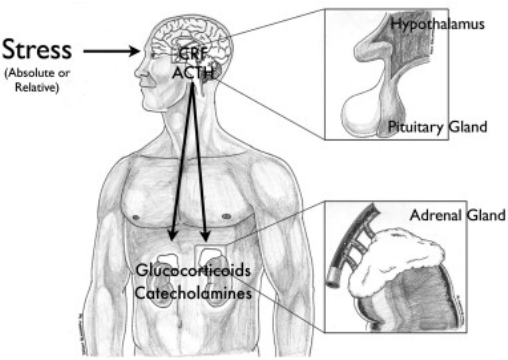}
\caption{Schematic representation of the hypothalamic-pituitary–adrenal (HPA) axis \cite{Lupien07BrainCog}.}\label{fig:gluco}
\end{figure}
\begin{figure}[!t]
\centering\includegraphics[scale=0.525]{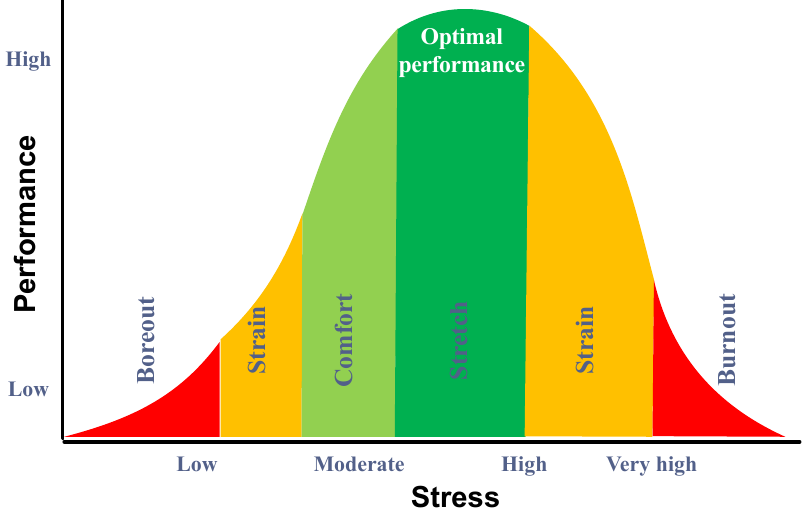}
\caption{Curvilinear relationship between stress and performance (from \protect\url{https://delphis.org.uk/peak-performance}).}\label{fig:bell}
\end{figure}

For staff, working is not merely a physical process but also a cognitive and emotional experience. When a situation (e.g., a high workload and a difficult task) is interpreted as being stressful, it triggers the activation of the hypothalamic–pituitary–adrenal axis whereby neurons in the hypothalamus release corticotropin-releasing hormone that subsequently triggers the release of adrenocorticotropin, which travels in the blood and reaches the adrenal glands above the kidneys to trigger the secretion of the so-called stress hormones (Fig. \ref{fig:gluco}) \cite{Lupien07BrainCog}. Psychologists have found that the relationship between emotional stress and working performance can be characterized by a bell-shaped curve (Fig. \ref{fig:bell}), where the performance is low for either very low or very high stress, and the best performance is achieved when the level of stress is just right. However, to the best of our knowledge, existing staff scheduling methods rarely consider work performance evolution, and hence often result in low-efficient or even impractical schedules due to biased estimates of working performance.

To overcome the above shortcomings, in this paper, we study a call-center staff scheduling problem that explicitly models the effect of emotional stress on working performance. First, we present an emotional stress driven working performance model, which estimates the working performance of employees based on skill levels and five emotional states (\emph{Depression}, \emph{Activation}, \emph{Anxiety}, \emph{Concentration}, and \emph{Endurance}), which are quantitatively assessed from personal characteristics and workload (working stress). On the basis of the model, we formulate a combined short-term and long-term call-center staff scheduling problem as a 0-1 programming model for maximizing the customer service level depending on staff working performance. To efficiently solve the problem, we propose a memetic optimization algorithm combining global mutation for evolving long-term schedules and deep reinforcement learning assisted variable neighborhood search for improving short-term schedules. We conduct computational experiments on real-world instances of call-center staff scheduling in the Zhejiang Branch of Industrial and Commercial Bank of China, and the results demonstrate the performance advantages of our methods over selected popular staff scheduling methods. In brief, the novel contributions of this paper is three-fold:
\begin{itemize}
\item The emotional stress driven working performance model;
\item The combined short- and long-term call-center staff scheduling problem based on the working performance model;
\item The memetic optimization algorithm for the scheduling problem.
\end{itemize}

In the rest of this paper, we review the related work in Section \ref{sec:liter}, present emotional stress driven working performance model in the Section \ref{sec:model}, and describe the problem formulation in Section \ref{sec:prob}. We then propose the memetic algorithm in Section \ref{sec:alg}, present the experimental results in Section \ref{sec:exp}, and finally concludes with discussions in Section \ref{sec:concl}.

\section{Related Work}\label{sec:liter}
Staff scheduling is a common problem to most organizations. This paper focuses on call-center staff scheduling, a main difficulty of which is that the number and features of tasks are not exactly known \textit{a priori} \cite{Ernst04EJOR}. In some studies \cite{Henderson99,Parkan99IJMS}, such information was predicted in the planning horizon using queuing models and simulation models. \"{O}rmeci et al. \cite{Ormeci14Omega} studied a staff rostering problem for a call center considering the transportation costs of agents from/to their homes, for which they developed a mixed integer programming model that incorporates the shuttle requirements at the beginning and end of the shifts into the agent-shift assignment decisions. T\"{u}rker and Demiriz \cite{Turker18MPE} studied shift scheduling and rostering in inbound call centers with overlapping shift systems; they developed both integer programming and constraint programming models, and their results showed that the constraint programming model was more efficient. For a bank information technologies staff scheduling problem, Labidi et al. \cite{Labidi14SWJ} developed a multiobjective programming model for minimizing the overtimes while maximize the satisfaction of employees. Kierm et al. \cite{Kierm16IIE} studied a more general flexible cyclic rostering problem that is common throughout the service industry; they formulated the problem as a multi-stage stochastic program and developed two approximation methods involving reductions to a two-stage problem.

Early studies typically formulated staff scheduling problems as linear programs, integer programs, or mixed ones, for which exact mathematical programming approaches can be directly applied. However, these exact approaches  can be very time-consuming,  formidable, or even impossible to solve large-scale real life problems that are often with millions or even billions variables to optimality \cite{Ernst04EJOR}. Heuristics and metaheuristics have been the favorite choices for dealing with these complex problems \cite{CaoZ24TCyb}. 
Specific heuristic methods for solving staff scheduling problems have been developed based on linear programming, backtracking search \cite{Heydrich20ORHC}, Lagrangian relaxation \cite{ZhouJ21TRPC}, branch-and-price \cite{FengT24TRPB}, etc. 
Compared to problem-specific heuristics, metaheuristics are simpler to implement and more general-purpose, allowing problem specific information to be incorporated and exploited. Dowling et al. \cite{Dowling97ANOR} adopted a simulated annealing (SA) algorithm in a system for rostering rosters airline ground staffs for a major international airline at one of the busiest international airports over a monthly planning horizon, which obtained higher quality solutions than genetic algorithm (GA) and tabu search. Ceschia et al. \cite{Ceschia23ORHC} used an SA based on a combination of two neighborhoods for nurse rostering, which was applied to a set of health facilities in northern Italy. To solve a problem of scheduling staff with mixed skills and under multi-criteria including cost and staff surplus, Cai and Li \cite{CaiX00EJOR} developed a new GA that improves the traditional GA in ranking-based selection, multi-point crossover, and a specific heuristic for infeasibility handling. In \cite{Aickelin00JSche}, Aickelin and Dowsland developed another improved GA for a nurse rostering problem, employing co-evolutionary subpopulations to exploit the structure of the problem constraints. Chen et al. \cite{ChenJ22CIE} combined a GA with a workforce assignment heuristic to solve an integrated resource-constrained multi-project scheduling and multi-skilled workforce assignment problem. Akjiratikarl et al. \cite{Akjir07CIE} adapted the continuous particle swarm optimization (PSO) to a home care worker scheduling problem that is discrete in nature, using a heuristic assignment scheme to transform the continuous vectors to job schedules. Todorovic and Petrovic \cite{Todoro13TSys} used a bee colony optimization algorithm for nurse rostering by alternating the constructive phase to assign  shifts to nurses and the local search phase to improve the solution quality. Lu et al. \cite{Lu23IJPR} adapted water wave optimization (WWO) to a bank customer service representative scheduling problem, using the principles and guidelines of the metaheuristic for combinatorial optimization \cite{Zheng19ASOC}. Soukour et al. \cite{Soukour13ESWA} studied a staff scheduling problem in airport security service, where staff assignment is performed using a memetic algorithm combining dedicated encoding, crossover, and  three neighborhood functions. There are also researchers using multi-objective evolutionary algorithms to obtain the Pareto optimal set of multi-objective scheduling/rostering problems \cite{ChenC17TSys,ZhangZ21TCyb,ZhouS21TITS}, which are significantly more complex than their single-objective counterparts.

The existing approaches generally regarded staff as uniform and emotionless agents,  except considering different skill levels. Only a few studies considered physiological factors such as heart rate \cite{Tran23TSys} and fatigue \cite{ZhaoF25TCyb} or psychological feelings related to factors such as well-being \cite{Petrovic21PATAT}. Nevertheless, the effect of work schedules on emotional stress and the consequent effect on working performance are rarely take into account in the literature. This study aims to fill this gap: by explicitly modeling and incorporating emotional stress, our method reflects a more realistic understanding and utilization of human behavior in staff scheduling.

\section{Emotional Stress Driven Working Performance Model}\label{sec:model}
\subsection{Basic Law of Arousal Effects on Performance}
The present working performance model is based on the Yerkes-Dodson Law of arousal effects on memory performance \cite{Yerkes08}, which stated that there is a curvilinear relationship between arousal and performance:  strong emotionality can increase attention and interest to enhance performance, and impair performance under more complex or challenging situations, such as in divided attention, multitasking, and working memory tasks \cite{Diamond07NeuPlast}.

\begin{figure}[!t]
\centering\includegraphics[scale=0.5]{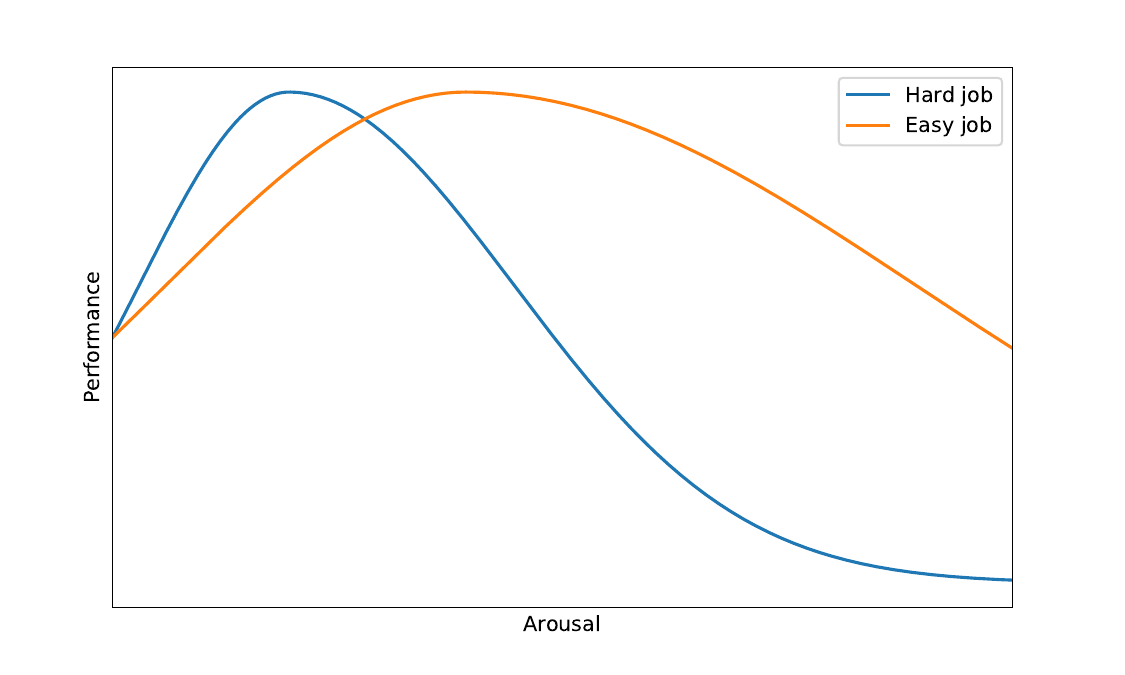}
\caption{Different arousal effects on performance under easy and hard situations.}\label{fig:hard-easy}
\end{figure}

Different effects observed on human behavioral performance are akin to the notion of the Yerkes-Dodson Law.  When performing complex tasks, a person's learning or working performance gradually increases with increasing glucocorticoid levels caused by stress, and the optimal state is achieved when Type I glucocorticoid receptors are saturated and there is partial occupancy of Type II glucocorticoid receptors; afterwards, the performance will deteriorate if the person is continuously exposure to stressful situations. Follow-up studies (e.g., \cite{Denenberg60,Anderson94PID,Dickman02HumanFac}) have also reinforced the work of Yerkes-Dodson by taking into account the difficulty of the task as an intervening variable in arousal effects on performance, showing that under more difficult or challenging tasks, the learning rate can be more efficient to approach the optimal state (as the onset of stress activates the hippocampus to produce a dramatic increase in levels of intracellular calcium), but then deteriorate more rapidly when high stress suppresses hippocampal functioning \cite{Diamond07NeuPlast} (Fig. \ref{fig:hard-easy}).
%

\subsection{Short-Term Working Performance}
\begin{figure}[!t]
\centering\includegraphics[scale=0.5]{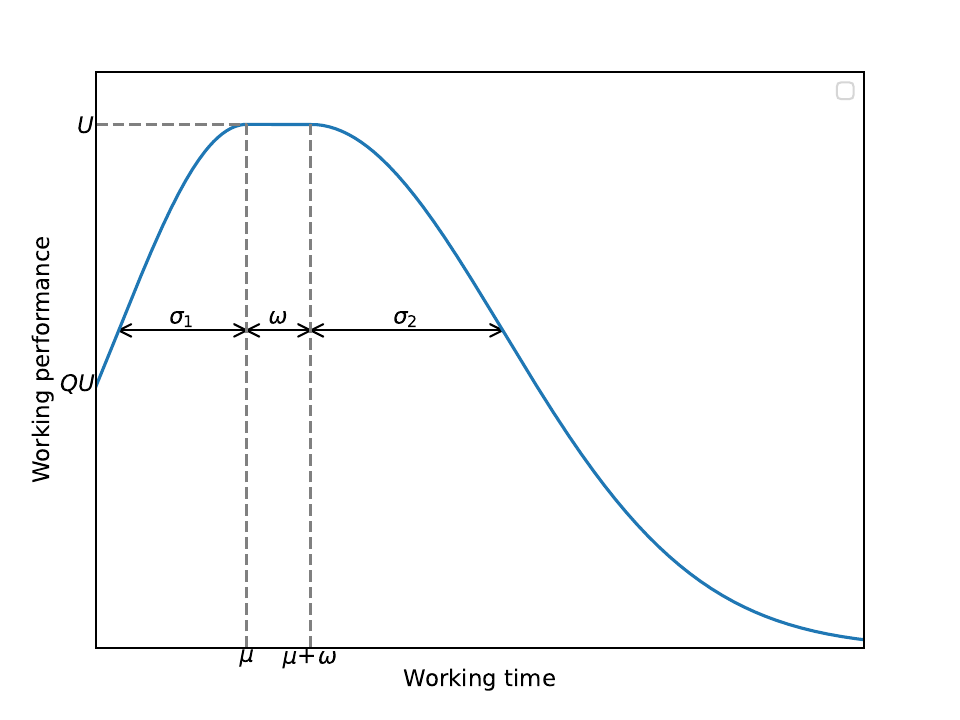}
\caption{Different arousal effects on performance under easy and hard situations.}\label{fig:perf-short}
\end{figure}

After beginning work after plenty of rest, an employee's working performance gradually increases to the optimal state, remains at the state for a short period, and then deteriorates with increasing working time, reflected by divided attention and decrease in reaction time. As shown in Fig. \ref{fig:perf-short}, we use two Gaussian functions with centers $\mu$ and $\mu+\omega$ and deviations $\sigma_1$ and $\sigma_2$ to depict the rising curve and the falling curve, respectively, which have the same maximum function value $U$ at the centers. The rising curve starts at time 0, at which the initial performance is $QU$ ($Q$ is an initial ratio), and uses a duration of $\mu$ to reach the maximum performance $U$. The peak period length is $\omega$. The falling curves starts at time $\mu+\omega$, and drops to approximately 60.6\%, 13.5\%, and 1.11\% of the maximum performance after duration of $\sigma_2$, $2\sigma_2$, and $3\sigma_2$, respectively.

The key parameters of the performance curve are related to an occupational state \emph{Skill\_level} (valued in the range of [0,1]) and five psychological/emotional states including \emph{Depression}, \emph{Activation}, \emph{Anxiety}, \emph{Concentration}, and \emph{Endurance} (all leveled in the range from 1 to 5) considered in this study as follows.

\begin{itemize}
\item The maximum performance $U$  is positively related to \emph{Skill\_level}  and negatively related to \emph{Depression}:
	\begin{equation}\small
	U= \textit{Skill\_level} \times \frac{6-\textit{Depression}}{5}
	\label{eq:max_perf}\end{equation}
\item The initial performance ratio $Q$ is positively related to \emph{Activation}:
	\begin{equation}\small
	Q= \big(\frac{\textit{Activation}}{6}\big)^{\frac{1}{2}}
	\end{equation}
\item The performance rising length $\sigma_1$ is negatively related to \emph{Activation} and positively related to \emph{Anxiety}:
	\begin{equation}\small
	\sigma_1= T_\text{r}\big(\frac{11-\textit{Activation}}{5}\big)^{1+\frac{\textit{Anxiety}}{5}}
	\end{equation}
	where $ T_\text{r}$ is a baseline performance rising time length.
\item The performance peak period length $\omega$ is positively related to \emph{Concentration} and negatively related to \emph{Anxiety}:
	\begin{equation}\small
	\omega= T_\text{p}\big(1+\frac{\textit{Concentration}}{5}\big) \big(\frac{6-\textit{Anxiety}}{5}\big)
	\end{equation}
	where $ T_\text{p}$ is a baseline performance peak length.
\item The performance falling length $\sigma_2$ is positively related to \emph{Endurance} and negatively related to \emph{Anxiety}:
	\begin{equation}\small
	\sigma_2= T_\text{f}\big(1+\frac{\textit{Endurance}}{5}\big)^{\frac{7-\textit{Anxiety}}{2}}
	\label{eq:sigma2}\end{equation}
	where $ T_\text{f}$ is a baseline performance falling time length.
\end{itemize}

\subsection{Long-Term Working Performance}
The five emotional states aforementioned depend on personal/physiological characteristics, living situations, and working stress \cite{Chaudhary23PakJHSS}, the factors of which used in this study are listed in Table \ref{tab:factors}. Nevertheless, the collection of factors is open. For example, if available, more financial situation factors such as personal/family incomes and expenses, can be used to assess the economic stress more accurately; we use \emph{PIT} and \emph{PIT\_SAD} because they are easy to acquire and can reflect economic income and burden in a relatively comprehensive way.

\newcommand{\tabincell}[2]{\begin{tabular}{@{}#1@{}}#2\end{tabular}}
\begin{table*}
\caption{Factors used for the assessment of emotional states in this study \label{tab:factors}}
\centering
\begin{tabular}{|l||l|}\hline
Type & Factors\\\hline
Personal/physiological & \emph{Age}, \emph{Gender}, body mass index (\emph{BMI})\\
Living situation & \emph{Commuting\_hours}, \emph{Sleeping\_hours}\\
Financial situation & personal income tax (\emph{PIT}), PIT special additional deduction (\emph{PIT\_SAD})\\\hline
Working stress & \tabincell{l}{years of employment (\emph{YOE}), continuous working days $D^\text{C}$, working days $D^\text{M}$ and $D^\text{S}$ in the last 30 and 90 days, respectively,\\
	 working hours $T^\text{D}$ in the current day, working hours $T^\text{W}$, $T^\text{M}$, and $T^\text{S}$ in the last 7, 30, and 90 days, respectively, \\
	 hard working hours $T^\text{D\dag}$ in the current day, hard working hours $T^\text{W\dag}$, $T^\text{M\dag}$, and $T^\text{S\dag}$ in the last 7, 30, and 90 days, respectively.} \\\hline
\end{tabular}
\end{table*}

A psychological test is conducted (normally during induction training) to grade the basic level of each emotional state, denoted by \emph{Depression$_0$}, \emph{Activation$_0$}, \emph{Anxiety$_0$}, \emph{Concentration$_0$}, and \emph{Endurance$_0$}. Taking the basic levels and the 19 factors as inputs, a neuro-fuzzy network (called Emotion NFN) is constructed to assess each of the five emotional states via TSK fuzzy rules \cite{Takagi85TSMC} with the following form:
\begin{align}
& \small\textnormal{IF } (x_1 \textnormal{ is } A^r_1) \textnormal{ AND } (x_2 \textnormal{ is } A^r_2) \ldots \textnormal{ AND } (x_p \textnormal{ is } A^r_p) \nonumber\\
& \small\textnormal{THEN } y \textnormal{ is } g^r(\mathbf{x}) \label{eq:fuzz-rule}
\end{align}
where $p\!=\!20$ is the number of inputs, $x_i$ is the $i$-th input, $A^r_i$ is the fuzzy set of $x_i$ in the current rule $r$ ($1\!\le\! i\!\le\! p$), and $g^r(\mathbf{x})$ is a linear combination in the form of $c^r_0+c^r_1 x_1+\ldots+c^r_p x_p$. 

The network consists of four layers for fuzzy inference:
\begin{itemize}
\item Layer 1 (\emph{the membership layer}),  where each node calculates the membership $\mu_{A^r_i}(x_i)$ of each $i$-th input to each $r$-th rule.
\item Layer 2 (\emph{the rule layer}), where each node calculates the firing strength of each $r$-th rule as a t-norm of all membership values $\mu_{A^r_i}(x_i)$ as
	\begin{equation}\small
	f^r(\mathbf{x}) = \prod_{i=1}^{p}\mu_{A^r_i}(x_i)
	\label{eq:firing}\end{equation}
\item Layer 3 (\emph{the weighting layer}), which normalizes the firing strengths as:
	\begin{equation}\small
	w_r = \frac{f^r(\mathbf{x})}{\sum_{r=1}^{R}f^r(\mathbf{x})}
	\label{eq:norm_strength}\end{equation}
\item Layer 4 (\emph{the output layer}), which calculates a center of sets defuzzifier \cite{Sugeno93TFS} that combines the outputs of all rules:
	\begin{equation}\small
	\hat{g}(\mathbf{x}) =\sum_{r=1}^{R}w_r g^r(\mathbf{x})
	\label{eq:defuzz}\end{equation}
\end{itemize}

In addition, another neuro-fuzzy network (called Impairment NFN) is established to assess the degree of monolithic impairment of performance because of high levels of stress (posttraumatic stress disorder) \cite{Lupien07BrainCog,Jelles21JPsyRes} based on the five emotional states: if the impairment degree is assessed as \emph{Severe}, the person is considered as not suitable to continue working and should receive medical (psychological) treatment.

Compared to black-box style neural networks, the neuro-fuzzy networks possess a good linguistically interpretability provided by fuzzy rules \cite{Mendel21TFS}. We employ a stochastic initialization and a fast online sequential training algorithm to configure the networks \cite{Rong09TSMCB,WangD17TCyb}.

\section{Combined Short- and Long-Term Customer Service Scheduling Optimization Problem}\label{sec:prob}
The considered problem is to schedule $m$ call-center employees from both short- and long-term perspectives. From the long-term perspective, it determines the working days and rest days of each employee in a decision period of $D$ days ($D\!=\!90$ in this study). From the short-term perspective, it determines the working hours and rest hours of and, for each working hour, whether easy or hard customer service requests are assigned to each employee everyday. According to their difficulties, all service requests are divided into the easy class and hard class. Moreover, each service request falls into one of the $N$ business categories. An assumption is that in each working hour, an employee only processes one class of requests in one business category, because switching categories back and forth will usually significantly affect the working performance. Therefore, the decision variables are represented by binary variables $x_{i,j,l,h,d}$ ($1\!\le\! i\!\le\! m; 1\!\le\! j\!\le\! N; 1\!\le\! l\!\le\! 2; 1\!\le\! h\!\le\!24; 1\!\le\! d\!\le\! D$, $l\!=\!1$ denotes easy class and $l\!=\!2$ denotes hard class), where values $x_{i,j,l,h,d}\!=\!0$ denotes that the $i$-th employee is allocated with the $l$-th class of jobs in the $j$-th business category in the $h$-th hour of the $d$-th day and $x_{i,j,l,h,d}\!=\!1$ otherwise.

Up to the beginning of the decision period, for each employee, the factors of working days and hours, as well as other factors in Table \ref{tab:factors} are known as input parameters, from which we use the Emotion NFN to assess the five emotional states; based on these states together with the skill level of the employee, we estimate the performance parameters according to Eqs. (\ref{eq:max_perf})--(\ref{eq:sigma2}), and thus estimate the working performance  $\textit{pf}_{i,j,l}$. Let $\Delta\tau_{j,l}$ denote the average time for processing a request of the $l$-th class in the $j$-th business category, the expected time duration for the employee to process a request is estimated as $\textit{pf}_{i,j,l} \Delta\tau_{j,l}$. 

Based on history data, the organization predicts the arrival rate $\lambda_{j,l,h,d}$ of customer service requests of each $l$-th class and $j$-th category in the $h$-th hour of $d$-th day. From the beginning time $t=0$, for each hour and each request type (class/category), we simulate the arrival of every new request after an interval $(-\ln u)/\lambda_{j,l,h,d}$,  where $u$ is a random number uniformly distributed in (0,1); whenever a new request arrives, if there are idle employees, the one with the maximum idle time is selected to process it; otherwise, the request enters into the waiting queue to wait an available employee.

\begin{figure}[!t]
\centering\includegraphics[scale=0.5]{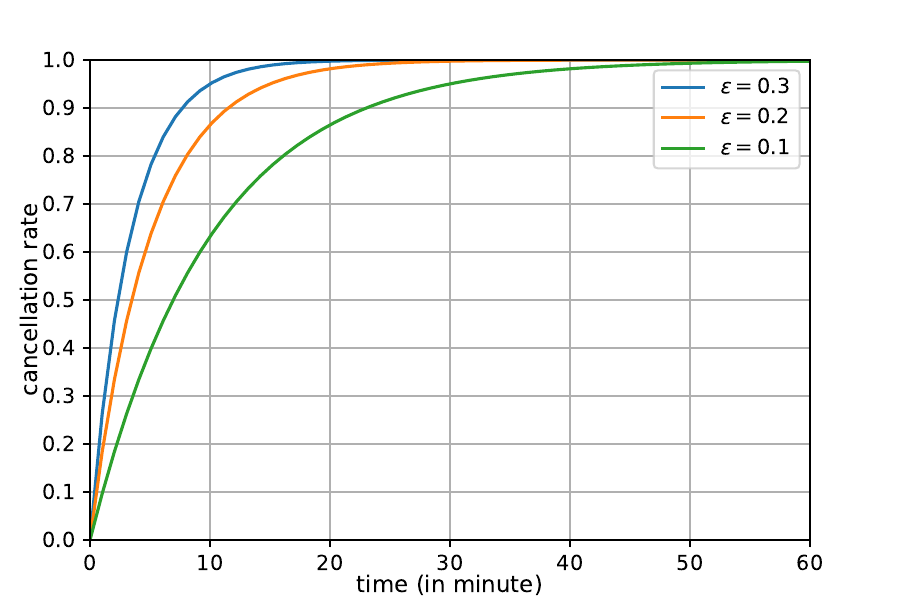}
\caption{Service request cancellation rate with time.}\label{fig:cancel}
\end{figure}

If the waiting time of a request exceeds a predefined threshold $\widehat{t}^\text{wait}_{j,l}$, it is regarded as a \emph{delayed} job. Moreover, a customer may cancel the request during waiting, and the cancellation probability increases with the waiting time. We calculate the cancellation probability of a request of the $l$-th class and $j$-th category as Eq. (\ref{eq:cancel}), where $\epsilon_{j,l}$ is the parameter related to the patience of customers in the corresponding group (the curves of which are shown in Fig. \ref{fig:cancel}):
\begin{equation}\small
p^\text{cancel}_{j,l}(t) = 1-\exp(-\epsilon_{j,l} t)
\label{eq:cancel}\end{equation}

Since a request enters into the waiting queue, every minute we generate a random number $u$ uniformly distributed in (0,1), and set it as canceled if $u<p^\text{cancel}_{j,l}(t)$.

\begin{figure*}
\centering\includegraphics[scale=0.64]{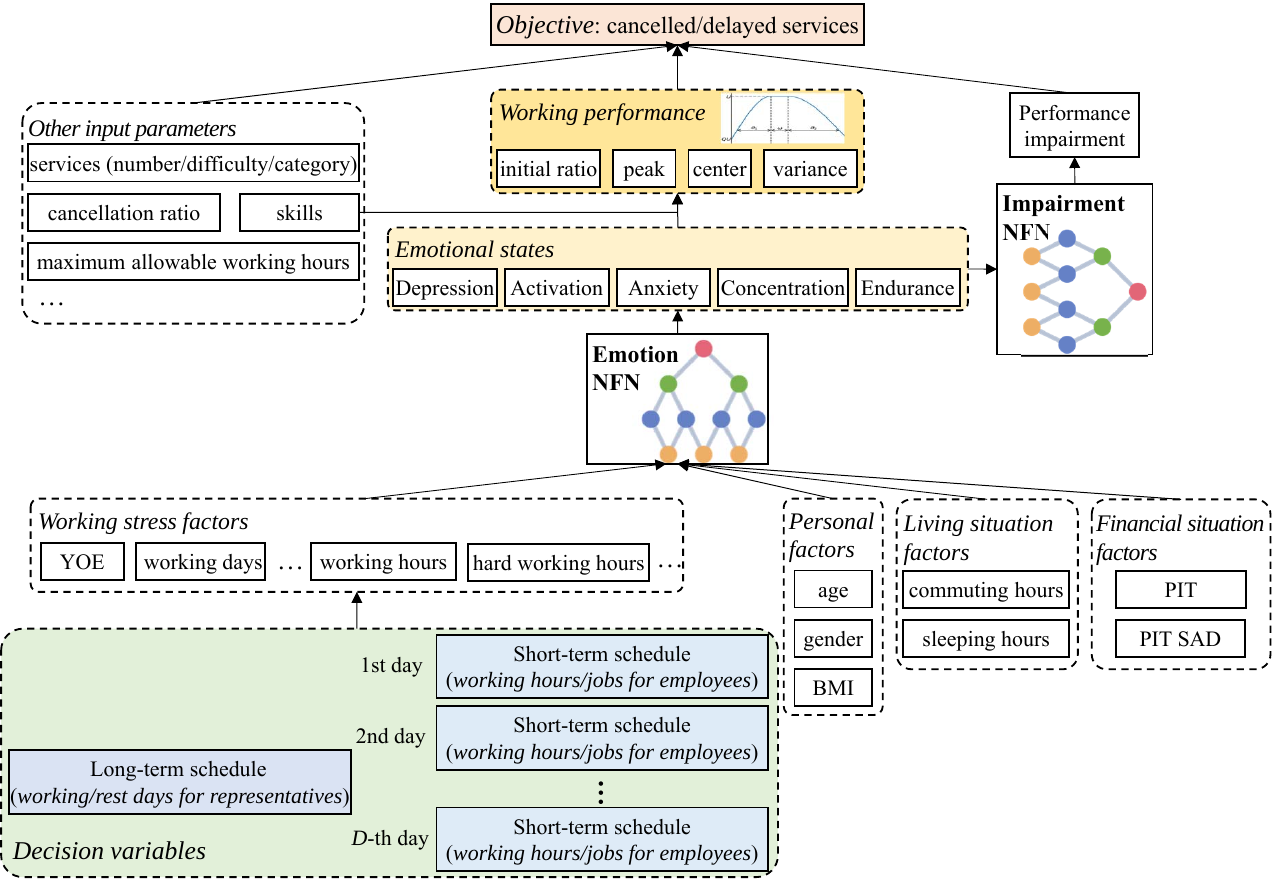}
\caption{Illustration of the structure of the call-center staff scheduling problem.}\label{fig:prob}
\end{figure*}

Consequently, we obtain the number $n_{j,l,h,d}$ of service requests, number $n^\text{delay}_{j,l,h,d}$ of delayed jobs, and number $n^\text{cancel}_{j,l,h,d}$ of canceled jobs of each type in each hour, as well as the set of requests that are still in the waiting queue at the end of current hour (and left to the next hour). After each hour, according to the job allocation and processing, we also update the working days/hours for each employee, use the Emotion NFN to re-assess the five emotional states, and then use the Impairment NFN to assess whether the employee has a severe impairment of performance. If not, we re-estimate the working performance, which is used in the simulation and calculation for the next hour. The underlying structure of the problem formulation is illustrated by Fig. \ref{fig:prob}.

Given the importance weight $w_{j,l}$ of each $l$-th class of service requests in each $j$-th business category, the problem aims to maximize the customer satisfaction by minimizing the aggregation of the following two metrics.
\begin{itemize}
\item The total weighted number of delayed service requests:
	\begin{equation}\small
	f^\text{delay}(X)= \sum_{j=1}^{N}\sum_{l=1}^{2}\sum_{d=1}^{D}\sum_{h=1}^{24}w_{j,l}n^\text{delay}_{j,l,h,d}
	\end{equation}
\item The total weighted number of canceled service requests:
	\begin{equation}\small
	f^\text{cancel}(X)= \sum_{j=1}^{N}\sum_{l=1}^{2}\sum_{d=1}^{D}\sum_{h=1}^{24}w_{j,l}n^\text{cancel}_{j,l,h,d}
	\end{equation}
\end{itemize}

The problem objective is defined as minimizing the weight aggregation of delay and cancellation as Eqs. (\ref{eq:obj}) (where $\omega$ is a positive weight larger than 1), subject to that each employee can be assigned with at most one type of job in each hour  (Eq. \ref{eq:constr_hour}), the working hours of each employee do not exceed the maximum daily working hours $\widehat{H}^\text{D}$ and maximum monthly working hours $\widehat{H}^\text{M}$ permitted by the labor law (Eq. \ref{eq:constr_daily} and Eq. \ref{eq:constr_month}), and no one will have a severe performance impairment caused by working stress (Eq. \ref{eq:constr_impair}).

\begin{figure*}[ht]
\begin{small}
\begin{align}
\min & f(X)= f^\text{delay}(X)+ \omega f^\text{cancel}(X) \label{eq:obj}\\
\text{s.t. } & \sum_{j=1}^{N}\sum_{l=1}^{2}x_{i,j,l,h,d}\le 1, \quad \forall 1\!\le\! i\!\le\! m; 1\!\le\!h\!\le\! 24; 1\!\le\! d\!\le\! D \label{eq:constr_hour}\\
  	&\sum_{j=1}^{N}\sum_{l=1}^{2}\sum_{h=1}^{24}x_{i,j,l,h,d}\le \widehat{H}^\text{D}, \quad \forall 1\!\le\! i\!\le\! m; 1\!\le\! d\!\le\! D \label{eq:constr_daily}\\
	&  \sum_{j=1}^{N}\sum_{l=1}^{2}\sum_{h=1}^{24}\sum_{d'=d}^{d\!+\!30} x_{i,j,l,h,d'}\le \widehat{H}^\text{M}, \quad \forall 1\!\le\! i\!\le\! m; 1\!\le\! d\!\le\! D-30 \label{eq:constr_month}\\
	&\textit{Impairment}_{i,h,d}\text{ is not } \textit{Severe}, \quad \forall 1\!\le\! i\!\le\! m; 1\!\le\! h\!\le\!24; 1\!\le\! d\!\le\! D \label{eq:constr_impair}\\
	& x_{i,j,l,h,d}\in \{0,1\}, \quad \forall 1\!\le\! i\!\le\! m; 1\!\le\! j\!\le\! N; 1\!\le\! l\!\le\! 2; 1\!\le\! h\!\le\!24; 1\!\le\! d\!\le\! D \nonumber
\end{align}
\end{small}
\end{figure*}

\section{Memetic Optimization Algorithm}\label{sec:alg}
\begin{figure}[!t]
\centering\includegraphics[scale=0.64]{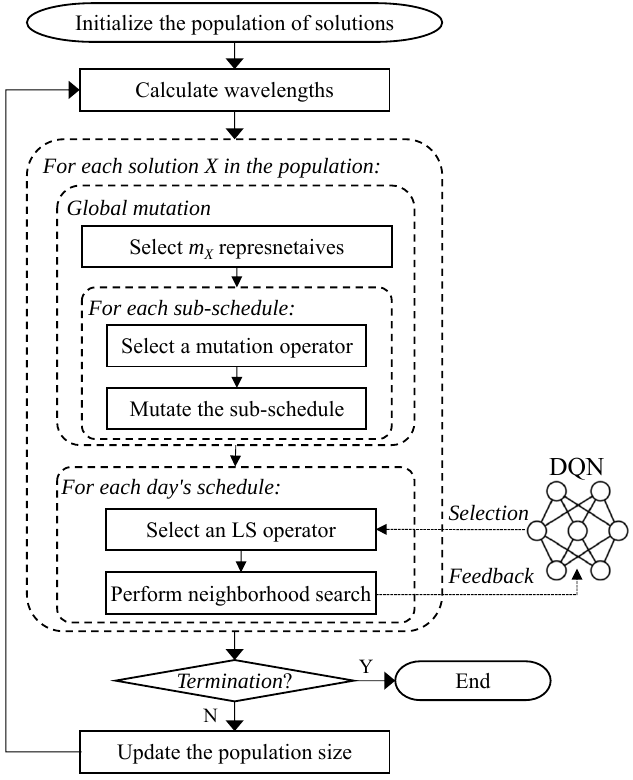}
\caption{Basic flowchart of the proposed memetic algorithm.}\label{fig:flow}
\end{figure}

The proposed memetic algorithm evolves a population of solutions to the problem by integrating global mutation for evolving long-term schedules with variable neighborhood search for improving short-term schedules, where neighborhood search operators are adaptively selected using reinforcement learning. Fig. \ref{fig:flow} presents the basic flowchart of the algorithm.

\subsection{Solution Initialization}\label{sec:alg_init}

To avoid random generation of many infeasible or low-effective solutions, we design the following procedure to initialize each solution $X$ considering the balance of work loads, importance of jobs, and the problem constraints:
\begin{enumerate}
\item For each day, calculate the number $n_{j,l,d}\!=\! \sum_{l=1}^2\sum_{h=1}^{24} n_{j,l,h,d}$ of service requests of each $l$-th class and $j$-th category, and then calculate the expected total time $\tau_{j,l,d}\!=\! n_{j,l,d}\Delta\tau_{j,l}$ for processing these jobs;
\item Based on the average daily working hours $\overline{H}^\text{D}$ (set to eight hours in this study), evaluate a basic required number $\overline{m}_{j,l,d}= \tau_{j,l,d}/\overline{H}^\text{D}$ of employees for processing these jobs;
\item For each type of job in non-increasing order of the importance weight $w_{j,l}$, generate a Gaussian number $m_{j,l,d}$ with center $\overline{m}_{j,l,d}$ and variance $\overline{m}_{j,l,d}/4$, and then randomly select  $m_{j,l,d}$ employees to process these jobs: we first try to select those employees with working performance higher than 0.5 and working hours in the current day shorter than $\overline{H}^\text{D}\!-\! \tau_{j,l,d}/\overline{m}_{j,l,d}$; if not sufficient, we continue to select those with working performance higher than 0.25 and working hours satisfying the problem constraints (\ref{eq:constr_daily}) and  (\ref{eq:constr_month}); if still not sufficient, we allow the jobs to be unprocessed;
\item Randomly assign the working hours $\tau_{j,l,d}$ to the selected employees, subject to that ratio of the maximum working hours to the minimum maximum working hours among the employees does not exceed 2.
\end{enumerate}

The first three steps generate a long-term schedule that determines the working days and rest days for all employees, and the fourth step generates $D$ short-term schedules, each arranging the working hours and job types to the selected employees in a day. Both long-term and short-term schedules are evolved by the global mutation and variable neighborhood search described in the following two subsections.

\subsection{Global Mutation for Long-Term Scheduling}\label{sec:alg_global}
The global mutation is conducted on the long-term schedule of a solution $X$ by randomly selecting a number $m_X$ of employees, and changing the sub-schedule of each selected employee in one of the following four ways.
\begin{enumerate}[MUT1]
\item (Add a rest day): If the number of working days of the employee is in the top 2/3 of all $m$ employees, randomly change a working day to a rest day, and then allocate the corresponding jobs to one or two employee whose working days are in the bottom 2/3.
\item (Add a working day): If the number of working days of the employee is in the bottom 2/3 of all $m$ employees, randomly add a working day for the employee, and then reallocate the working hours of that day according to Step 4) of  solution initialization procedure in subsection \ref{sec:alg_init}.
\item (Swap a working day and a rest day): Randomly change a working day to a rest day and change a rest day to a working day, and then reallocate the working hours of the two days according to Step 4) of  solution initialization procedure in subsection \ref{sec:alg_init}.
\item (Swap two working days between two employees): Randomly swap a working day of the employee with another working day of another employee (given that the working day of one employee is just the rest day of another employee), and then reallocate the working hours of the two days according to Step 4) of solution initialization procedure in subsection \ref{sec:alg_init}.
\end{enumerate}

The number $m_X$, which determines the mutation degree of the solution $X$, depends on the solution fitness. We use $\Lambda_X$ to denote the maximum the portion of the employees to be rescheduled, which is initially set to 0.5 for each solution, and then updated at each generation as
\begin{equation}\small
\Lambda_X= \Lambda_X \alpha^{-(f_{\max}-f(X)+\epsilon)/(f_{\max}-f_{\min}+\epsilon)}
\label{eq:wavelength}\end{equation}
where $f_{\max}$ and $f_{\min}$ are the maximum and minimum objective function values among the population, respectively, $\alpha$ is a reduction coefficient which is suggested to be 1.0026, and $\epsilon$ is a very small number to avoid division by zero. $m_X$ is set to the product of $M$ and a random value between 0 and $\Lambda_X$. In this way, low-fitness solutions (with large objective function values) have probably large portions to mutate and hence explore in large spaces, while high-fitness solutions have probably small portions to mutate and hence exploit in small regions, and thus the algorithm can achieve a good dynamic balance between diversification and intensification. This is exactly the optimization principle of the WWO metaheuristic \cite{Zheng15COR}, where $\Lambda_X$ is analogous to the ``wavelength'' of the solution in exploring the solution space.

\subsection{Variable Neighborhood Search for Short-Term Scheduling}\label{sec:alg_local}
Neighborhood search is performed on a solution $X$ by trying to improve the short-term schedule of each day using one of the following four operators.
\begin{enumerate}[LS1]
\item Reassign a working hour from an employee to another employee who is rest in that hour.
\item Swap the jobs between two employees (who have different job types) in the same hour.
\item Move a working hour of an employee to an earlier hour $h$, given that $\widetilde{n}_{j,l,h,d}\ge n_{i,j,l,h,d}/2$, i.e., the number of unprocessed service requests in the earlier hour is at least the half of the number of service requests the employee can process.
\item Move a working hour of an employee to a later hour $h$, given that $\widetilde{n}_{j,l,h,d}\ge n_{i,j,l,h,d}/2$, and the number of unprocessed service requests in the original hour after moving does not exceed a predefined threshold (set to 15 in this study).
\end{enumerate}

On each day's short-term schedule, the number of times the neighborhood search conducted is $N_\text{S}$; therefore, the total number of neighborhood solutions generated is $DN_\text{S}$, and the best one among them, if better than the original solution $X$, will replace $X$ in the population. The value of $N_\text{S}$ increases from a lower limit $N^\text{L}_\text{S}$ to an upper limit $ N^\text{U}_\text{S}$ with generation number $g$ as follows (where $g_{\max}$ denotes the maximum generation number of the algorithm), such that neighborhood search is increasingly enhanced in later stages:
\begin{equation}\small
N_\text{S}(g)= N^\text{L}_\text{S}+\big(\frac{g}{g_{\max}}\big)^2( N^\text{U}_\text{S}- N^\text{L}_\text{S})
\label{eq:nls}\end{equation}

\subsection{Deep Reinforcement Learning for Neighborhood Search}\label{sec:alg_dqn}
We utilize a deep Q-network (DQN) that combines deep neural network and Q-learning \cite{Mnih15Nature} to select among LS1--LS4 for each neighborhood search operation. Given a target solution $X$, for each operator LS$_i$ ($1\!\le\! i\!\le\! 4$), the DQN estimates a Q-value $Q(S_{t+1},\mathrm{LS}_i)$ as the conditional selection probability in each state $S$ defined by a quadruple $\{f(X)/f_\text{best}, \Delta f(X)/f_\text{best}, \textit{rank}, \textit{idl}\}$, where $f_\text{best}$ is the best objective function value found so far, $\Delta f(X)$ is the decrement of the objective function value of $X$ from the previous generation, $\textit{rank}$ is the rank of $X$ in the population, and $\textit{idl}$ is the index of the operator applied to $X$ at the last time. Taking the four state variables as inputs, the DQN uses two hidden layers with the ReLU activation function to perform forward calculation and finally outputs the conditional probabilities based on softmax activation \cite{Mnih15Nature}.


After a neighborhood search operator LS is selected according to the conditional selection probability and applied to the solution $X$, the solution is transferred to a new state, and the reward $r$ is calculated as $\Delta f/f_\text{best}$ if $f(X)$ is decreased (the fitness is improved) and is 0 otherwise. The DQN parameters $\mathbf{\theta}$ is trained using the Adam optimizer \cite{King15Adam} to minimize the loss function as follows (where $\gamma$ is the discount factor):
\begin{equation}\small
L= \mathbb{E} \big(r+ \gamma\max_i Q(S_{t+1}, \mathrm{LS}_i)- Q(S_t, \mathrm{LS})\big)^2
\label{eq:DQN-loss}\end{equation}

\subsection{Constraint Handling}\label{sec:alg-repair}
For each newly generated solution, we check for each employee whether the constraints (\ref{eq:constr_daily})--(\ref{eq:constr_impair}) are violated. If so, from the working hour at which the constraints just become violated, we change the working hours to rest hours one by one in a backward fashion until the constraints are satisfied. Then, we use the following procedure to deal with the jobs left by the employee day by day in a forward fashion:
\begin{enumerate}
\item For each working hour released from the employee in the day, from all other employees that are working in that day while vacant in that hour, select the one with the lowest workload in that day to undertake the left jobs in that hour without further violating the constraints;
\item If Step 1) fails to transfer all working hours in the day, from all other employees that are rest in that day, select the one with the lowest workload during the decision period to undertake the left jobs in that day without further violating the constraints; 
\item If Step 2) still fails, leave the jobs that are not transferred in Step 1) to increase the waiting times of jobs in the waiting queue.
\end{enumerate}

\subsection{Algorithm Framework}
Algorithm \ref{alg} presents the pseudo-code of the memetic algorithm. It also uses a population size reduction policy as Eq. (\ref{eq:pop-size}) (where $N^\text{U}_\text{P}$ and $N^\text{L}_\text{P}$ are the upper and lower limits of the population size, respectively) that facilitates global exploration in earlier stages and enhances local exploitation in later stages. 
\begin{equation} \small
N_\text{P}(g)= N^\text{U}_\text{P}-\big(\frac{g}{g_{\max}}\big)^2( N^\text{U}_\text{P}- N^\text{L}_\text{P})
\label{eq:pop-size}\end{equation}

Whenever the size should be reduced by one, among the solutions whose fitness are in the second half of the population, the one that has stayed (without improvement) in the population for the maximum number of generation is removed. 

\begin{algorithm}\small
\caption{Memetic algorithm for the problem. \label{alg}} 
	initialize a population of $N_\text{P}$ solutions\;
	\While{the termination condition is not satisfied} {
		\ForEach{solution $X$ in the population}{
			Calculate $\Lambda_X$ according to Eq. (\ref{eq:wavelength})\;
		}
		\ForEach{solution $X$ in the population}{
			\For{$i=1$ to $M$} {
				\If{$\textit{rand}(0,1)<\Lambda_X$}{
					Randomly select a mutation operator\;
					Perform mutation on the $i$-th subschedule\;
				}
			}
			\For{$d=1$ to $D$}{ 
				Use the DQN to choose a neighborhood search operator\;
				Perform neighborhood search on the $d$-th short-term schedule\;
			}
			\If{Constraints are violated}{
				Try to repair the solution according to the procedure in subsection \ref{sec:alg-repair}\;
			} 
		}
		Update the conditional  selection probabilities of the neighborhood search operators in DQN\;
		Calculate the population size according to Eq. (\ref{eq:pop-size})\;
		\If{the size is reduced}{
			Remove the most stagnant solution whose fitness is in the second half from the population\;
		}
	}
	\Return the best solution $X^*$ found so far.
\end{algorithm}

\section{Experiments}\label{sec:exp}
We first test the effectiveness of the emotional stress driven working performance model, and then test the scheduling performance of the memetic optimization algorithm.

\subsection{Experiments on Working Performance Assessment}
Working performance is stochastic in nature and depends on various of factors, many of which (e.g., sadness and anger) are difficult to observe and measure. Due to the limited number of factors observed, accurate prediction of working performance can be generally impractical. Therefore, our purpose is to validate the \emph{effectiveness} (rather than \emph{accuracy}) of the present model by comparing with several commonly used prediction models including the multiple linear regression (MLR), back-propagation artificial neural network (ANN), and adaptive neuro-fuzzy inference system  (ANFIS) \cite{Jang93} rather than those state-of-the-art complex models.

\begin{figure*}[!t]
\centering
\subfloat[]{\includegraphics[scale=0.64]{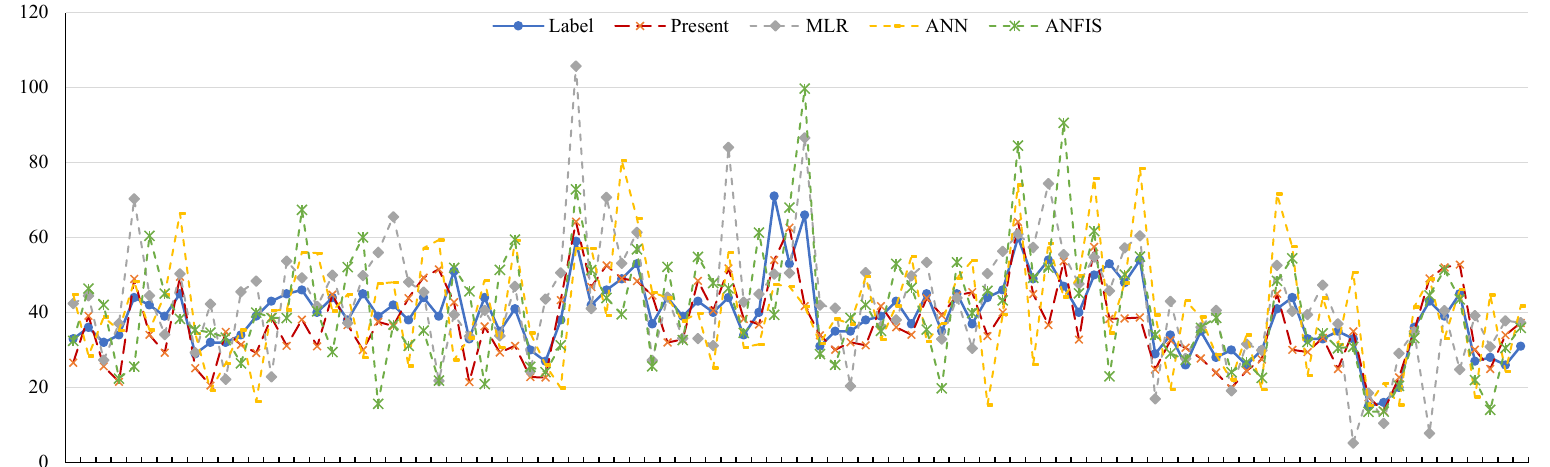} \label{fig:distrib-sig1}} \hfil
\subfloat[]{\includegraphics[scale=0.64]{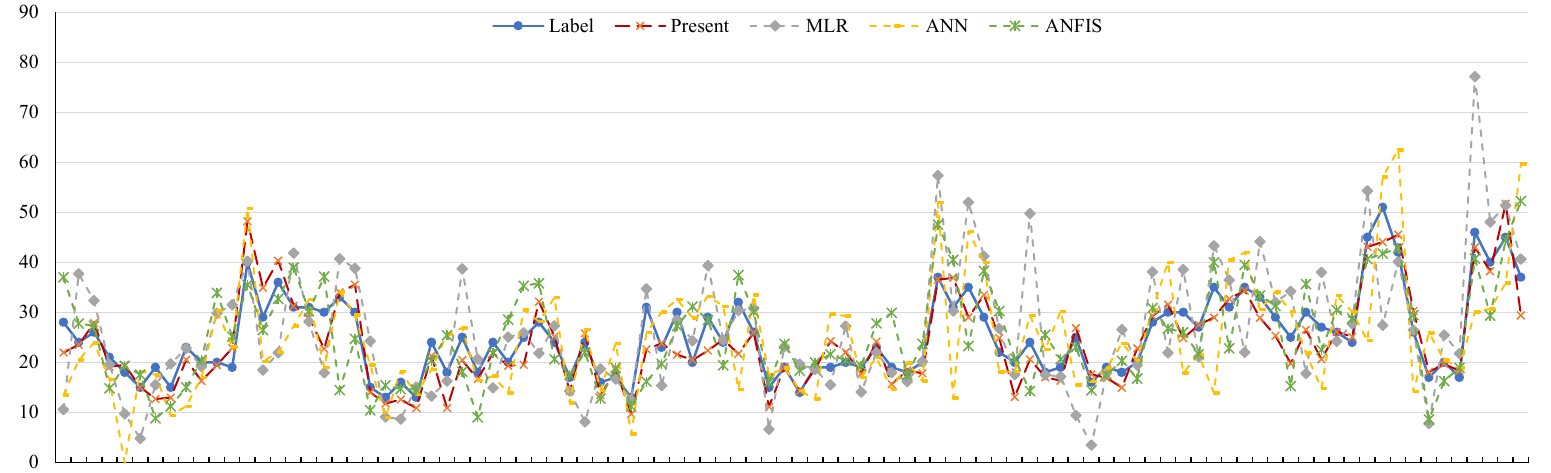} \label{fig:distrib-omega}} \hfil
\subfloat[]{\includegraphics[scale=0.64]{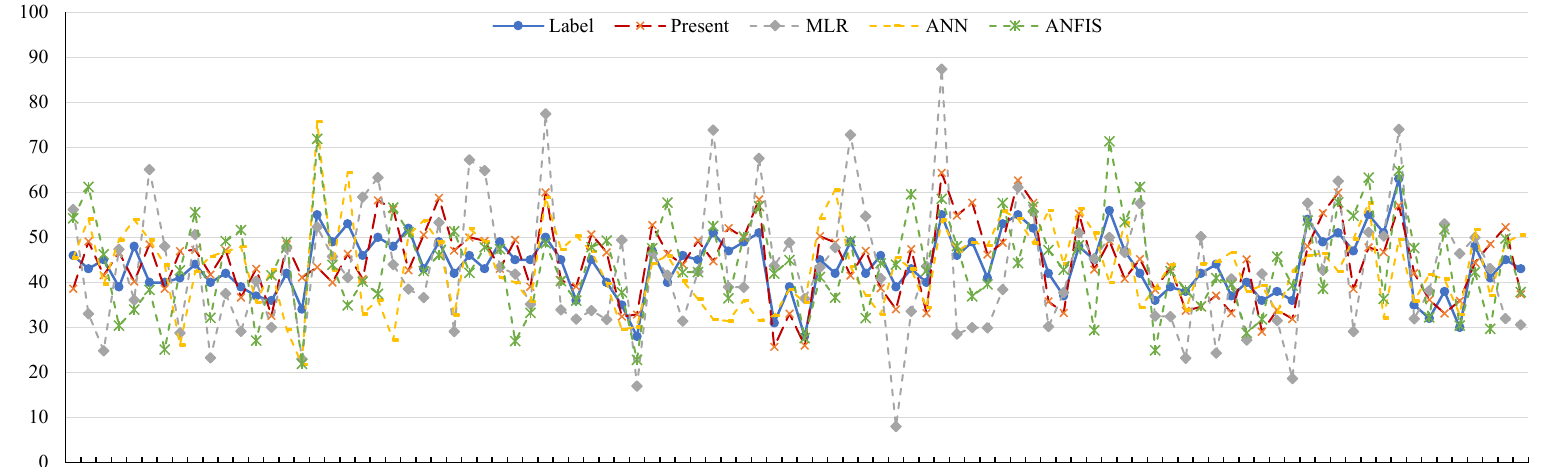} \label{fig:distrib-sig2}}
\caption{Working performance metric labels and values of the obtained by the four models. (a) length parameter $\sigma_1$ of the rising curve; (b) peak period length $\omega$; (c) length parameter $\sigma_2$ of the falling curve.}
\label{fig:distrib}
\end{figure*}

We test the models on a set of 102 samples for assessing three working performance metrics:  the length parameter $\sigma_1$ of the rising curve,  peak period length $\omega$, and length parameter $\sigma_2$ of the falling curve (see Fig. \ref{fig:perf-short}). Fig. \ref{fig:distrib}(a)--(c) presents the labels and the values obtained by the models, and Table \ref{tab:sig1}, Table \ref{tab:omega}, and Table \ref{tab:sig2} present the mean absolute error (MAE), mean squared error (MSE), explained variance score (EVS), and R2 score of the results obtained by the models for the three working performance metrics, respectively.

\begin{table}
\caption{Regression errors of the four models for the length parameter $\sigma_1$ of the rising curve. \label{tab:sig1}}
\centering
\begin{tabular}{|l|rrrr|}\hline
model &\multicolumn{1}{c}{MAE} &\multicolumn{1}{c}{MSE} &\multicolumn{1}{c}{EVC} &\multicolumn{1}{c|}{R$^2$}\\\hline
Present & \textbf{6.493792} & \textbf{61.266236} & \textbf{ 0.349747} & \textbf{0.287334} \\
MLR & 9.388516 & 157.298156 &  \underline{$\!-\!$0.699425} & \underline{$\!-\!$0.829735} \\
ANN & 9.909641 & 154.498058 &  \underline{$\!-\!$0.783561} & \underline{$\!-\!$0.797164} \\
ANFIS & 8.633993 & 137.831720 & \underline{$\!-\!$0.592193} & \underline{$\!-\!$0.603297} \\\hline
\end{tabular}
\end{table}

\begin{table}
\caption{Regression errors of the four models for thepeak period length $\omega$. \label{tab:omega}}
\centering
\begin{tabular}{|l|rrrr|}\hline
model &\multicolumn{1}{c}{MAE} &\multicolumn{1}{c}{MSE} &\multicolumn{1}{c}{EVC} &\multicolumn{1}{c|}{R$^2$}\\\hline
Present &  \textbf{2.990397} & \textbf{14.504770} & \textbf{0.788807} & \textbf{0.773619} \\
MLR  & 6.882661 & 81.747739  & \underline{$\!-\!$0.260725} & \underline{$\!-\!$0.275867} \\
ANN  & 6.703164 & 72.266339  & \underline{$\!-\!$0.123821} &  \underline{$\!-\!$0.127888} \\
ANFIS &  4.982674 & 39.015085 & 0.393318 & 0.391077 \\\hline
\end{tabular}
\end{table}

\begin{table}
\caption{Regression errors of the four models for the length parameter $\sigma_2$ of the falling curve. \label{tab:sig2}}
\centering
\begin{tabular}{|l|rrrr|}\hline
model &\multicolumn{1}{c}{MAE} &\multicolumn{1}{c}{MSE} &\multicolumn{1}{c}{EVC} &\multicolumn{1}{c|}{R$^2$}\\\hline
Present & \textbf{5.744362} & \textbf{37.311463}  & \textbf{0.104737} & \textbf{0.094509} \\
MLR  & 10.010993 & 146.450270 & \underline{$\!-\!$2.538741 } & \underline{$\!-\!$2.554119 } \\
ANN  & 6.825305 & 72.991709  & \underline{$\!-\!$0.770997 } &  \underline{$\!-\!$0.771395 } \\
ANFIS & 6.856364 & 73.039521 & \underline{$\!-\!$0.772405 } &  \underline{$\!-\!$0.772555 }  \\\hline
\end{tabular}
\end{table}

For assessing the rising curve of the working performance, our model obtains an MAE smaller than three (in minutes, and the mean absolute percentage error is approximately 16\%). This error, which seems to be not so small, is generally acceptable for this difficult prediction problem. In contrast,  the results of the other three models are not acceptable: their percentage errors are larger than 20\%, and their EVS and R2 scores are all negative, which indicates that the models are far from explaining the variance in the data set (worse than a baseline model predicting the mean of the ground truth).

The errors of the models become smaller for assessing the performance peak period length. Nevertheless, the EVS and R2 scores obtained by the MLR and ANN are still negative, indicating very poor fits with the samples. The results obtained by our model and the ANFIS are generally acceptable, and the error of our model is significantly smaller than that of ANFIS.

For assessing the performance falling curve, all the models perform poorly. Comparatively, our model obtains significantly lower errors than the other threes. Moreover, only our model obtains positive EVS and R2, which indicates that it is the only fit model for this very difficult prediction problem.

In summary, our model obtains significantly better results than all other models for assessing each of the working performance metrics. For the other three models, except that the results of ANFIS for assessing the performance peak period length are generally acceptable, the results of these comparative models are all very poor. The experimental results demonstrate that our model is the only one acceptable model for working performance assessment. This is mainly because we appropriately identify the five emotional states and establish the hierarchical relationships for estimating the emotional states from the limited observed factors and then assessing the working performance from the emotional states. The other models, including the MLR that directly estimates the working performance from the observed factors and ANN and ANFIS that automatically infer one or two middle layers to perform a stepwise estimation of the working performance, fail to effectively model the relationship between the input factors and the output working performance. That is, the five emotional states play a critical role in effective working performance assessment.

\subsection{Experiments on Staff Scheduling}
We use five problem instances of call-center staff scheduling in the Zhejiang Branch of Industrial and Commercial Bank of China. Table \ref{tab:par} gives other main parameters of the problem (same for all  instances); Table \ref{tab:ins} presents the different characteristics of the instances.

\begin{table}[h]
\caption{Main parameters of the problem, where the last two columns denote values for easy and hard jobs, respectively.  \label{tab:par}}
\centering
\begin{tabular}{|l||rr|}\hline
Parameter & Easy & Hard \\\hline
$T_\text{r}$ & 45 min & 25 min \\
$T_\text{p}$ & 24 min & 18 min  \\
$T_\text{f}$ & 72 min & 36 min \\
$\widehat{H}^\text{D}$ & \multicolumn{2}{c|}{10 h} \\
$\widehat{H}^\text{M}$ & \multicolumn{2}{c|}{212 h} \\
$\omega$ & \multicolumn{2}{c|}{2} \\\hline
\end{tabular}
\end{table}

\begin{table}[h]
\caption{Main characteristics of the five instances, where $\overline{n}_{j,l,h,d}$ is the average number of services requests in a hour, $\overline{\textit{pf}}_{i,j,l}(0)$ the initial average performance of the employees, and $\overline{\epsilon}_{j,l}$ the average cancellation coefficient in Eq. (\ref{eq:cancel}). The values in a pair denote $l\!=\!1$ and $2$, respectively.  \label{tab:ins}}
\centering
\begin{tabular}{|l||cccccc|}\hline
Instance & $m$ & $N$ & $\overline{n}_{j,l,h,d}$ & $\overline{\textit{pf}}_{i,j,l}(0)$ & $\overline{\epsilon}_{j,l}$ & \\\hline
\#1 & 327 & 6 & (1535, 620) & (0.75, 0.73) & (0.20, 0.18) & \\
\#2 & 341 & 6 & (1712, 643) & (0.80, 0.71) & (0.20, 0.18) &  \\
\#3 & 383 & 8 & (2246, 855) & (0.72, 0.64) & (0.23, 0.20) & \\
\#4 & 396 & 8 & (2439, 849) & (0.69, 0.67) & (0.23, 0.20) & \\
\#5 & 442 & 9 & (2728, 877) & (0.68, 0.65) & (0.24, 0.18) & \\\hline
\end{tabular}
\end{table}

The proposed memetic algorithm combined with WWO-based mutation and DQN-assisted neighborhood search, denoted by MA-DQN, is compared with the following six methods from the literature (which are adapted to our problem to optimize the objective function in Eq. \ref{eq:obj}):
\begin{itemize}
\item The memetic algorithm for multi-skill staff scheduling \cite{Soukour13ESWA}, denoted by Shift-MA, which encodes a solution as an assignment of shifts to employees on the decision horizon, and then solves the problem in three steps including days-off scheduling, shift scheduling, and staff assignment, using combined GA and local search. 
\item A GA \cite{Rattan16} using the basic 0-1 encoding and genetic crossover and mutation, while utilizing a penalty function to handle the constraints.
\item A multi-thread simulated annealing (MTSA) algorithm enhanced with the multi-neighborhood \cite{Turan21IJPR}.
\item A hybrid GA and linear programming (GA-LP) algorithm \cite{YouP21SciProg}, using GA for working days allocations and LP for job scheduling in each day.
\item A hybrid GA and simulated annealing (GA-SA) algorithm \cite{Barghi22ORF}, using GA for working days allocations and SA for job scheduling in each day.
\item A WWO algorithm \cite{Lu23IJPR} using the basic 0-1 encoding and discretized operators.
\end{itemize}

All these six comparative algorithms do not consider working performance variance under emotional stress, i.e., regarding the working performance as constant values. In addition, we implement two variants of our proposed algorithm: the first also does not consider working performance variance under emotional stress, denoted by MA-DQN-NE; the second further removes DQN, denoted by MA-NE, using a random policy for selecting neighborhood search operators.

The key control parameters of the proposed algorithm are set as $N^\text{U}_\text{P}\!=\! 60$, $N^\text{L}_\text{P}\!=\! 10$,  $N^\text{U}_\text{S}\!=\! 45$, $N^\text{L}_\text{S}\!=\! 10$, and $\gamma\!=\!0.85$. The key parameters of the other methods from the literature are also tuned on the test set. For a fair comparison, we set the maximum number of fitness evaluation to $600m$ as the same termination condition for all algorithms. On each instance, each algorithm is run for 30 times, and a nonparametric Wilcoxon rank sum test is conducted on the result of MA-DQN and the result of each other algorithm on each instance.

\begin{figure}
\centering\includegraphics[scale=0.48]{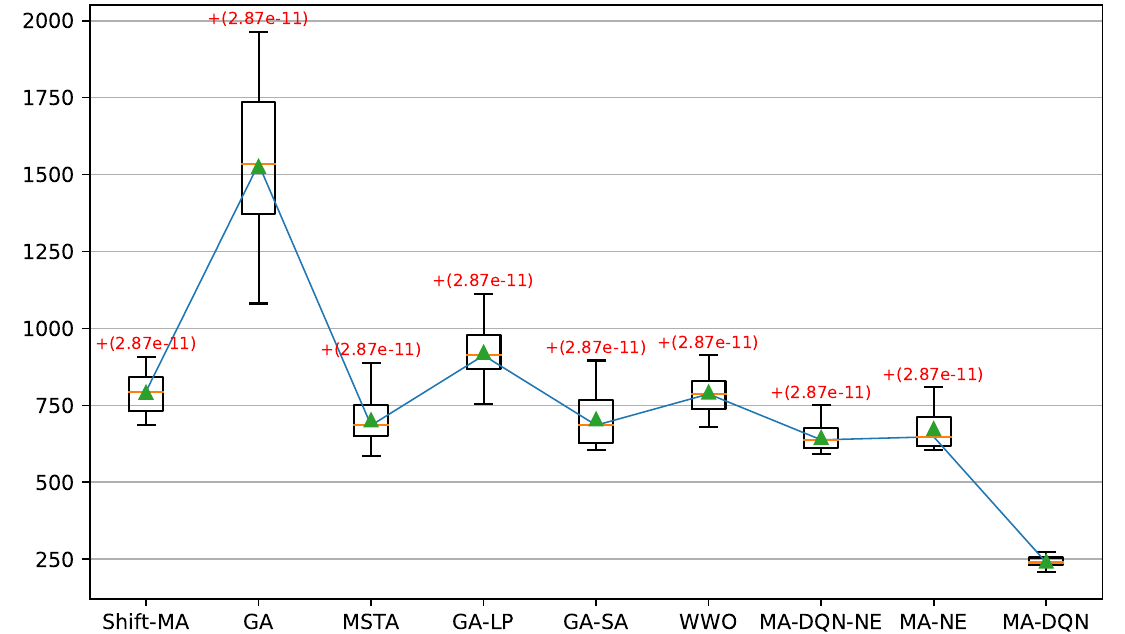}
\caption{Box plots of the results obtained by the nine algorithms on the first problem instance.}\label{fig:res1}
\end{figure}

\begin{figure}
\centering\includegraphics[scale=0.48]{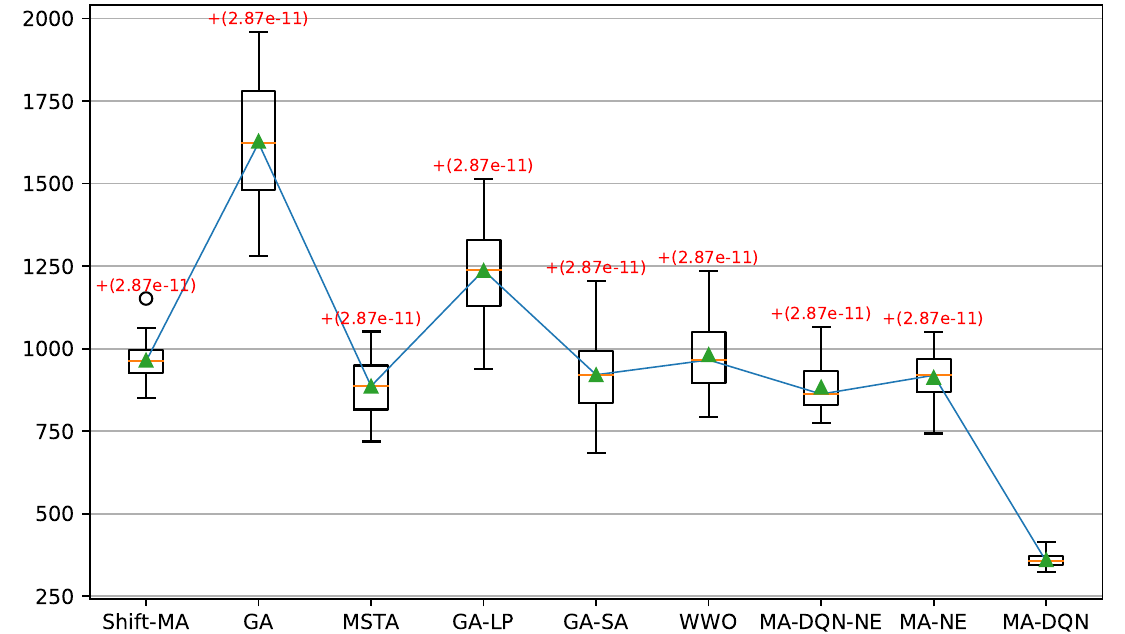}
\caption{Box plots of the results obtained by the nine algorithms on the second problem instance.}\label{fig:res2}
\end{figure}

\begin{figure}
\centering\includegraphics[scale=0.48]{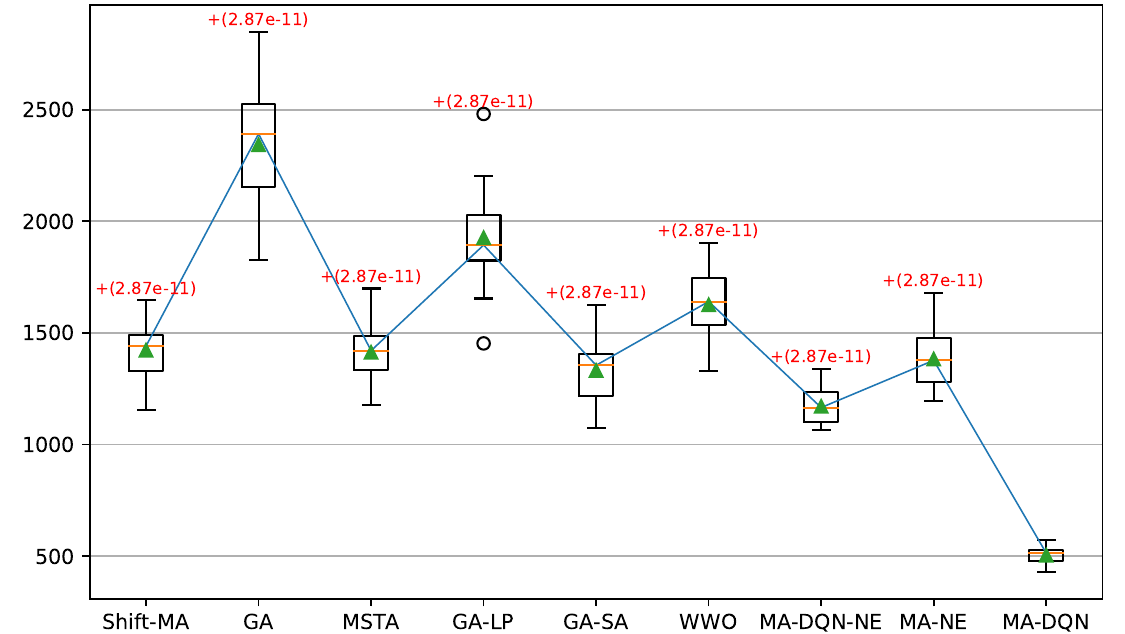}
\caption{Box plots of the results obtained by the nine algorithms on the third problem instance.}\label{fig:res3}
\end{figure}

\begin{figure}
\centering\includegraphics[scale=0.48]{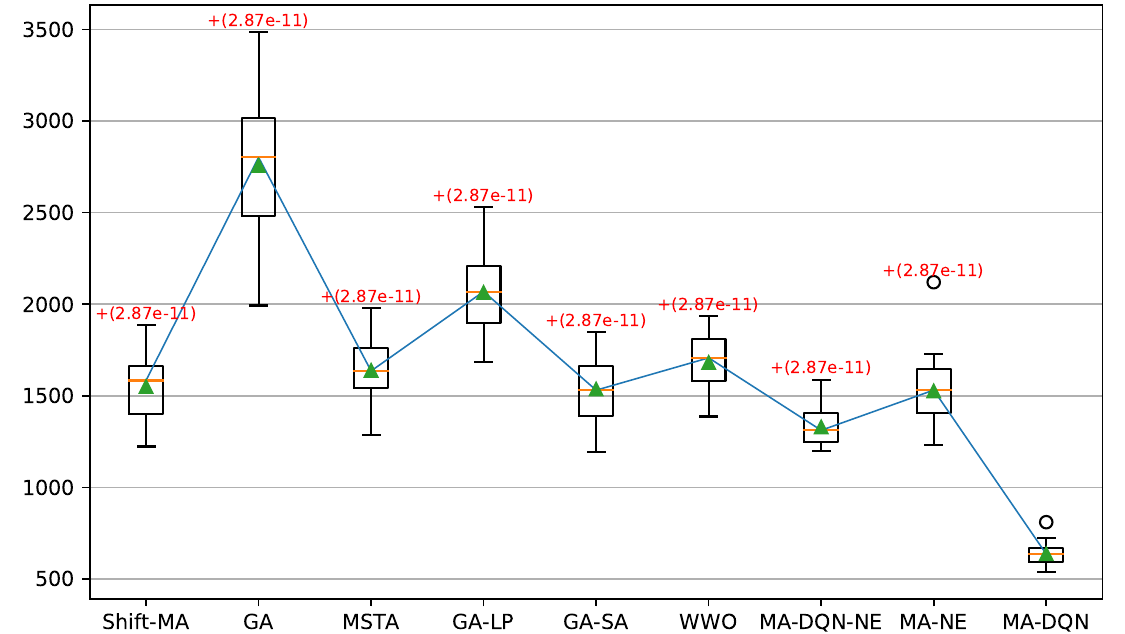}
\caption{Box plots of the results obtained by the nine algorithms on the fourth problem instance.}\label{fig:res4}
\end{figure}

\begin{figure}
\centering\includegraphics[scale=0.48]{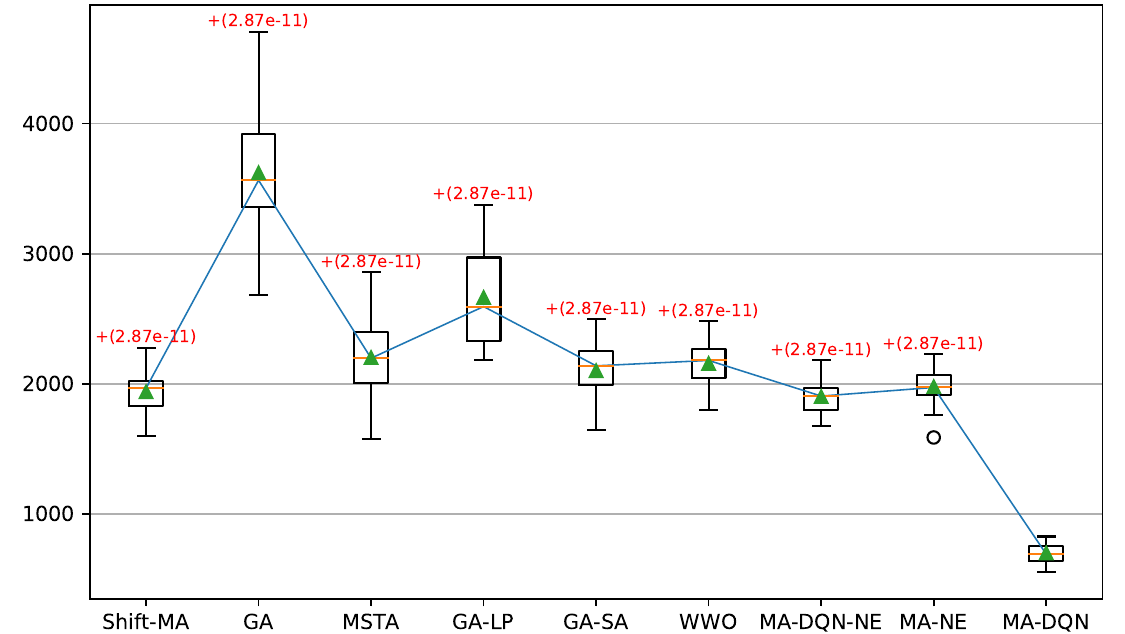}
\caption{Box plots of the results obtained by the nine algorithms on the fifth problem instance.}\label{fig:res5}
\end{figure}

The box plots in Figs. \ref{fig:res1}--\ref{fig:res5} present the distributions of the objective function values, including the mean (shown in green triangle) median, maximum, minimum, the first quartile (Q1) and the third quartile (Q3), obtained by the nine algorithms on the five test instances, respectively. Any objective function value below the lower limit $\text{Q}1\!-\!1.5(\text{Q}3\!-\!\text{Q}1)$ or above the upper limit $\text{Q}3\!+\!1.5(\text{Q}3\!-\!\text{Q}1)$ is drawn as an outlier. The number above the maximum line of the box of each other algorithm is the $p$-value of the corresponding rank sum test, and a red mark `+' before the $p$-value indicates there is a statistically significant difference at a confidence level of $95\%$ ($p\!<\!0.05$).

Among the comparative algorithms, the GA performs the worst on all test instances, mainly because both the basic 0-1 encoding and and crossover operator have two deficiencies: (1) limited diversity that often leads to early convergence; (2) the lack of fine-tuning ability. GA-LP and GA-SA employ linear programming and SA to make up the second deficiency, respectively, and both achieve performance improvement over the basic GA; the improvement of GA-SA is more significant than that of GA-LP, demonstrating that the SA heuristic is more efficient in daily job scheduling. However, the first deficiency of the GA still remains.

The performance of WWO is between GA-LP and GA-SA. The basic WWO balances the global and local based on its built-in propagation and breaking operators, which, as demonstrated by the results, is more efficient in solving this problem than GA-LP. However, as the breaking operator is relatively random and simple, its local search ability is weaker than the dedicated designed SA. The performance of MA-NE, which combines WWO-based mutation and random neighborhood search, reaches or exceeds that of GA-SA. MA-NE also outperforms Shift-MA on most test instances, demonstrating that WWO-based mutation provides better global search ability than GA crossover and mutation.

MA-DQN-NE obtains significantly better performance than MA-NE, validating the effectiveness of the deep reinforcement learning for adaptive selection of neighborhood search operators compared to the random search policy. Actually, the performance of MA-DQN-NE is the best among the eight algorithms without considering the effects of emotional stress, demonstrating the superiority of the proposed memetic algorithm designed for the  scheduling problem.

By taking  the effects of emotional stress on working performance into consideration, MA-DQN obtains the best result that is significantly better than all other algorithms on each test instance. The median objective function values of MA-DQN are only around 15\%$\sim$20\% of those of the the worst performing GA and are around 30\%$\sim$40\% of those of the other comparative algorithms. Particularly, the median objective function values of MA-DQN are 36.3\%$\sim$48.5\% of those of MA-DQN-NE, which clearly shows the significant improvement of scheduling efficiency brought by the consideration of emotional stress: without this consideration, the assumed working performance used in the other algorithms can deviate significantly from the actual working performance, and the consequent inaccurate evaluations of the objective function can make the algorithms unable to find the appropriate search areas in the solution space. The objective function values obtained by MA-DQN also have the smallest standard deviation among all nine algorithms, demonstrating that the consideration of emotional stress also reduces the uncertainty of function evaluation and therefore improves the robustness of the results.

\section{Conclusion}\label{sec:concl}
This paper presents an emotional stress driven model for quantitatively assessing the working performance of employees based on skill levels and emotional states, where emotional states are estimated from personal characteristics and workload. On the basis of the model, we formulate a combined short- and long-term call-center staff scheduling problem to maximize the customer service level. To efficiently solve the problem, we propose a memetic optimization algorithm combining global mutation and neighborhood search assisted by deep reinforcement learning. Experimental results on the real-world instances demonstrate the performance advantages of the proposed method.

Ongoing work is extending the working performance model for staff performing emergency tasks in harsh environments, e.g., firefighters and emergency medical technicians, whose emotional states play a more important role in their missions. The extended models can then be used for those emergency scheduling problems, which are critical for saving lives.



\begin{thebibliography}{10}
\providecommand{\url}[1]{#1}
\csname url@samestyle\endcsname
\providecommand{\newblock}{\relax}
\providecommand{\bibinfo}[2]{#2}
\providecommand{\BIBentrySTDinterwordspacing}{\spaceskip=0pt\relax}
\providecommand{\BIBentryALTinterwordstretchfactor}{4}
\providecommand{\BIBentryALTinterwordspacing}{\spaceskip=\fontdimen2\font plus
\BIBentryALTinterwordstretchfactor\fontdimen3\font minus
  \fontdimen4\font\relax}
\providecommand{\BIBforeignlanguage}[2]{{%
\expandafter\ifx\csname l@#1\endcsname\relax
\typeout{** WARNING: IEEEtran.bst: No hyphenation pattern has been}%
\typeout{** loaded for the language `#1'. Using the pattern for}%
\typeout{** the default language instead.}%
\else
\language=\csname l@#1\endcsname
\fi
#2}}
\providecommand{\BIBdecl}{\relax}
\BIBdecl

\bibitem{Ernst04EJOR}
A.~Ernst, H.~Jiang, M.~Krishnamoorthy, and D.~Sier, ``Staff scheduling and
  rostering: A review of applications, methods and models,'' \emph{Euro. J.
  Oper. Res.}, vol. 153, no.~1, pp. 3--27, 2004.

\bibitem{Nguyen22TCyb}
S.~Nguyen, D.~Thiruvady, M.~Zhang, and D.~Alahakoon, ``Automated design of
  multipass heuristics for resource-constrained job scheduling with
  self-competitive genetic programming,'' \emph{IEEE Trans. Cybern.}, vol.~52,
  no.~9, pp. 8603--8616, 2022.

\bibitem{Lupien07BrainCog}
S.~J. Lupien, F.~Maheu, M.~Tu, A.~Fiocco, and T.~E. Schramek, ``The effects of
  stress and stress hormones on human cognition: Implications for the field of
  brain and cognition,'' \emph{Brain Cognition}, vol.~65, no.~3, pp. 209--237,
  2007.

\bibitem{Henderson99}
S.~Henderson, A.~Mason, I.~Ziedins, and R.~Thomson, ``A heuristic for
  determining efficient staffing requirements for call centres,'' Department of
  Engineering Science, University of Auckland, Tech. Rep., 1999.

\bibitem{Parkan99IJMS}
C.~Parkan, K.~Lam, and H.~C. and, ``Workforce planning at a bill inquiries
  centre,'' \emph{Int. J. Model. Simul.}, vol.~19, no.~2, pp. 118--126, 1999.

\bibitem{Ormeci14Omega}
E.~L. \"{O}rmeci, F.~S. Salman, and E.~Y\"{u}cel, ``Staff rostering in call
  centers providing employee transportation,'' \emph{Omega}, vol.~43, pp.
  41--53, 2014.

\bibitem{Turker18MPE}
T.~T\"{u}rker and A.~Demiriz, ``An integrated approach for shift scheduling and
  rostering problems with break times for inbound call centers,'' \emph{Math.
  Prob. Eng.}, vol. 2018, no.~1, p. 7870849, 2018.

\bibitem{Labidi14SWJ}
M.~Labidi, M.~Mrad, A.~Gharbi, and M.~A. Louly, ``Scheduling {IT} staff at a
  bank: A mathematical programming approach,'' \emph{Sci. World J.}, vol. 2014,
  no.~1, p. 768374, 2014.

\bibitem{Kierm16IIE}
F.~Kiermaier, M.~Frey, and J.~F.~B. and, ``Flexible cyclic rostering in the
  service industry,'' \emph{IIE Trans.}, vol.~48, no.~12, pp. 1139--1155, 2016.

\bibitem{CaoZ24TCyb}
Z.~Cao, C.~Lin, M.~Zhou, and X.~Wen, ``Learning-based genetic algorithm to
  schedule an extended flexible job shop,'' \emph{IEEE Trans. Cybern.},
  vol.~54, no.~11, pp. 6909--6920, 2024.

\bibitem{Heydrich20ORHC}
S.~Heydrich, R.~Schroeder, and S.~Velten, ``Collaborative duty rostering in
  health care professions,'' \emph{Oper. Res. Health Care}, vol.~27, p. 100278,
  2020.

\bibitem{ZhouJ21TRPC}
J.~Zhou, X.~Xu, J.~Long, and J.~Ding, ``Integrated optimization approach to
  metro crew scheduling and rostering,'' \emph{Transp. Res. Part C: Emerg.
  Technol.}, vol. 123, p. 102975, 2021.

\bibitem{FengT24TRPB}
T.~Feng, R.~M. Lusby, Y.~Zhang, S.~Tao, B.~Zhang, and Q.~Peng, ``A
  branch-and-price algorithm for integrating urban rail crew scheduling and
  rostering problems,'' \emph{Transp. Res. Part B: Method.}, vol. 183, p.
  102941, 2024.

\bibitem{Dowling97ANOR}
D.~Dowling, M.~Krishnamoorthy, H.~Mackenzie, and D.~Sier, ``Staff rostering at
  a large international airport,'' \emph{Ann. Oper. Res.}, vol.~72, pp.
  125--147, 1997.

\bibitem{Ceschia23ORHC}
S.~Ceschia, L.~{Di Gaspero}, V.~Mazzaracchio, G.~Policante, and A.~Schaerf,
  ``Solving a real-world nurse rostering problem by simulated annealing,''
  \emph{Oper. Res. Health Care}, vol.~36, p. 100379, 2023.

\bibitem{CaiX00EJOR}
X.~Cai and K.~N. Li, ``A genetic algorithm for scheduling staff of mixed skills
  under multi-criteria,'' \emph{Euro. J. Oper. Res.}, vol. 125, no.~2, pp.
  359--369, 2000.

\bibitem{Aickelin00JSche}
U.~Aickelin and K.~A. Dowsland, ``Exploiting problem structure in a genetic
  algorithm approach to a nurse rostering problem,'' \emph{J. Scheduling},
  vol.~3, no.~3, pp. 139--153, 2000.

\bibitem{ChenJ22CIE}
J.~C. Chen, Y.-Y. Chen, T.-L. Chen, and Y.-H. Lin, ``Multi-project scheduling
  with multi-skilled workforce assignment considering uncertainty and learning
  effect for large-scale equipment manufacturer,'' \emph{Comput. Ind. Eng.},
  vol. 169, p. 108240, 2022.

\bibitem{Akjir07CIE}
C.~Akjiratikarl, P.~Yenradee, and P.~R. Drake, ``{PSO}-based algorithm for home
  care worker scheduling in the {UK},'' \emph{Comput. Ind. Eng.}, vol.~53,
  no.~4, pp. 559--583, 2007.

\bibitem{Todoro13TSys}
N.~Todorovic and S.~Petrovic, ``Bee colony optimization algorithm for nurse
  rostering,'' \emph{IEEE Trans. Syst. Man Cybern.: Syst.}, vol.~43, no.~2, pp.
  467--473, 2013.

\bibitem{Lu23IJPR}
X.~Lu, C.~Wu, X.~Yang, M.~Zhang, and Y.~Zheng, ``Adapted water wave
  optimization for integrated bank customer service representative
  scheduling,'' \emph{Int. J. Prod. Res.}, vol.~61, no.~1, pp. 320--335, 2023.

\bibitem{Zheng19ASOC}
Y.-J. Zheng, X.-Q. Lu, Y.-C. Du, Y.~Xue, and W.-G. Sheng, ``Water wave
  optimization for combinatorial optimization: Design strategies and
  applications,'' \emph{Appl. Soft Comput.}, vol.~83, p. 105611, 2019.

\bibitem{Soukour13ESWA}
A.~{Abdoul Soukour}, L.~Devendeville, C.~Lucet, and A.~Moukrim, ``A memetic
  algorithm for staff scheduling problem in airport security service,''
  \emph{Expert Syst. Appl.}, vol.~40, no.~18, pp. 7504--7512, 2013.

\bibitem{ChenC17TSys}
C.-H. Chen and J.-H. Chou, ``Multiobjective optimization of airline crew roster
  recovery problems under disruption conditions,'' \emph{IEEE Trans. Syst. Man
  Cybern.: Syst.}, vol.~47, no.~1, pp. 133--144, 2017.

\bibitem{ZhangZ21TCyb}
Z.-X. Zhang, W.-N. Chen, H.~Jin, and J.~Zhang, ``A preference biobjective
  evolutionary algorithm for the payment scheduling negotiation problem,''
  \emph{IEEE Trans. Cybern.}, vol.~51, no.~12, pp. 6105--6118, 2021.

\bibitem{ZhouS21TITS}
S.-Z. Zhou, Z.-H. Zhan, Z.-G. Chen, S.~Kwong, and J.~Zhang, ``A multi-objective
  ant colony system algorithm for airline crew rostering problem with fairness
  and satisfaction,'' \emph{IEEE Trans. Intell. Transp. Syst.}, vol.~22,
  no.~11, pp. 6784--6798, 2021.

\bibitem{Tran23TSys}
T.-A. Tran, M.~P\'{e}ntek, H.~Motahari-Nezhad, J.~Abonyi, L.~Kov\'{a}cs,
  L.~Gul\'{a}csi, G.~Eigner, Z.~Zrubka, and T.~Ruppert, ``Heart rate
  variability measurement to assess acute work-content-related stress of
  workers in industrial manufacturing environment--a systematic scoping
  review,'' \emph{IEEE Trans. Syst. Man Cybern.: Syst.}, vol.~53, no.~11, pp.
  6685--6692, 2023.

\bibitem{ZhaoF25TCyb}
F.~Zhao, H.~Zhou, L.~Wang, and Y.~Yu, ``A feature-based learning differential
  evolution algorithm for the flexible job-shop scheduling with occupational
  repetitive actions index,'' \emph{IEEE Trans. Cybern.}, vol.~55, no.~7, pp.
  3457--3470, 2025.

\bibitem{Petrovic21PATAT}
S.~Petrovic, J.~Parkin, and D.~Wrigley, ``Personnel scheduling considering
  employee well-being: insights from case studies,'' in \emph{Proc. 13th Int'l
  Conf. Practice and Theory of Automated Timetabling}, vol.~1, 2021, pp.
  10--23.

\bibitem{Yerkes08}
R.~M. Yerkes and J.~D. Dodson, ``The relation of strength of stimulus to
  rapidity of habit-formation,'' \emph{J. Compar. Neurol. Psychol.}, vol.~18,
  no.~5, pp. 459--482, 1908.

\bibitem{Diamond07NeuPlast}
D.~M. Diamond, A.~M. Campbell, C.~R. Park, J.~Halonen, and P.~R. Zoladz, ``The
  temporal dynamics model of emotional memory processing: A synthesis on the
  neurobiological basis of stress-induced amnesia, flashbulb and traumatic
  memories, and the yerkes-dodson law,'' \emph{Neural Plasticity}, vol. 2007,
  pp. 1--33, 2007.

\bibitem{Denenberg60}
V.~H. Denenberg and G.~G. Karas, ``Supplementary report: The {Yerkes-Dodson}
  law and shift in task difficulty,'' \emph{J. Exp. Psychol.}, vol.~59, no.~6,
  pp. 429--430, 1960.

\bibitem{Anderson94PID}
K.~J. Anderson, ``Impulsitivity, caffeine, and task difficulty: A
  within-subjects test of the {Yerkes-Dodson} law,'' \emph{Personality
  Individual Differences}, vol.~16, no.~6, pp. 813--829, 1994.

\bibitem{Dickman02HumanFac}
S.~J. Dickman, ``Dimensions of arousal: Wakefulness and vigor,'' \emph{Human
  Factors}, vol.~44, no.~3, pp. 429--442, 2002.

\bibitem{Chaudhary23PakJHSS}
S.~Chaudhary, N.~Nasir, S.~ur~Rahman, and S.~Masood~Sheikh, ``Impact of work
  load and stress in call center employees: Evidence from call center
  employees,'' \emph{Pakistan J. Humanit. Social Sci.}, vol.~11, no.~1, pp.
  160--171, 2023.

\bibitem{Takagi85TSMC}
T.~Takagi and M.~Sugeno, ``Fuzzy identification of systems and its applications
  to modeling and control,'' \emph{IEEE Trans. Syst. Man Cybern.}, vol.~15,
  no.~1, pp. 116--132, 1985.

\bibitem{Sugeno93TFS}
M.~Sugeno and T.~Yasukawa, ``A fuzzy-logic-based approach to qualitative
  modeling,'' \emph{IEEE Trans. Fuzzy Syst.}, vol.~1, no.~1, pp. 7--30, 1993.

\bibitem{Jelles21JPsyRes}
L.~Jellestad, N.~A. Vital, J.~Malamud, J.~Taeymans, and C.~Mueller-Pfeiffer,
  ``Functional impairment in posttraumatic stress disorder: A systematic review
  and meta-analysis,'' \emph{J. Psychiatric Res.}, vol. 136, pp. 14--22, 2021.

\bibitem{Mendel21TFS}
J.~M. Mendel and P.~P. Bonissone, ``Critical thinking about explainable {AI
  (XAI)} for rule-based fuzzy systems,'' \emph{IEEE Trans. Fuzzy Syst.},
  vol.~29, no.~12, pp. 3579--3593, 2021.

\bibitem{Rong09TSMCB}
H.-J. Rong, G.-B. Huang, N.~Sundararajan, and P.~Saratchandran, ``Online
  sequential fuzzy extreme learning machine for function approximation and
  classification problems,'' \emph{IEEE Trans. Syst. Man Cybern. Part B},
  vol.~39, no.~4, pp. 1067--1072, 2009.

\bibitem{WangD17TCyb}
D.~Wang and M.~Li, ``Stochastic configuration networks: Fundamentals and
  algorithms,'' \emph{IEEE Trans. Cybern.}, vol.~47, no.~10, pp. 3466--3479,
  2017.

\bibitem{Zheng15COR}
Y.-J. Zheng, ``Water wave optimization: A new nature-inspired metaheuristic,''
  \emph{Comput. Oper. Res.}, vol.~55, no.~1, pp. 1--11, 2015.

\bibitem{Mnih15Nature}
V.~Mnih, K.~Kavukcuoglu, D.~Silver, and {et al}, ``Human-level control through
  deep reinforcement learning,'' \emph{Nature}, vol. 518, pp. 529--533, 2015.

\bibitem{King15Adam}
\BIBentryALTinterwordspacing
D.~P. Kingma and J.~Ba, ``Adam: A method for stochastic optimization,'' in
  \emph{3rd International Conference for Learning Representations}, 2015.
  [Online]. Available: \url{http://arxiv.org/abs/1412.6980}
\BIBentrySTDinterwordspacing

\bibitem{Jang93}
J.~S.~R. Jang, ``{ANFIS}: Adaptive-network-based fuzzy inference system,''
  \emph{IEEE Trans. Syst. Man Cyber}, vol.~23, no.~3, pp. 665--685, 1993.

\bibitem{Rattan16}
T.~Rattanamanee, ``Multi-workday ergonomic workforce scheduling with personal
  and task constraint,'' Ph.D. dissertation, Thammasat University, 2016.

\bibitem{Turan21IJPR}
H.~H. Turan, F.~Kosanoglu, and M.~Atmis, ``A multi-skilled workforce
  optimisation in maintenance logistics networks by multi-thread simulated
  annealing algorithms,'' \emph{Int. J. Prod. Res.}, vol.~59, no.~9, pp.
  2624--2646, 2021.

\bibitem{YouP21SciProg}
P.-S. You and Y.-C. Hsieh, ``A heuristic algorithm for medical staff's
  scheduling problems with multiskills and vacation control,'' \emph{Sci.
  Progr.}, vol. 104, pp. 1--22, 2021.

\bibitem{Barghi22ORF}
B.~Barghi and S.~S. Sikari, ``Meta-heuristic solution with considering setup
  time for multi-skilled project scheduling problem,'' \emph{Oper. Res. Forum},
  vol.~16, no.~3, 2022.

\end{thebibliography}


%
%
%
%
%

\vfill

\end{document}